\documentclass[runningheads]{llncs}
\usepackage[T1]{fontenc}
\usepackage{graphicx}
\usepackage{amsmath,amsfonts,bm}
\usepackage{xcolor}
\RequirePackage[colorlinks=true,citecolor=blue,urlcolor=blue]{hyperref}

\def\eqref#1{equation~\ref{#1}}

\def\1{\bm{1}}

\DeclareMathAlphabet{\mathsfit}{\encodingdefault}{\sfdefault}{m}{sl}
\SetMathAlphabet{\mathsfit}{bold}{\encodingdefault}{\sfdefault}{bx}{n}

\newcommand{\E}{\mathbb{E}}

\newcommand{\R}{\mathbb{R}}

\newcommand{\Cov}{\mathrm{Cov}}

\usepackage{hyperref}
\usepackage{url}

\usepackage{booktabs}
\usepackage{amssymb}
\usepackage{caption}
\usepackage{subcaption}
\newcommand{\norm}[1]{\left\lVert#1\right\rVert}
\usepackage{multirow}
\newcommand{\B}{\boldsymbol}

\usepackage{pifont}
\usepackage{wrapfig}
\usepackage{paralist} 
\usepackage{enumitem}
\newcommand\eqdef{\mathrel{\overset{\makebox[0pt]{\mbox{\normalfont\tiny\sffamily def}}}{=}}}
\DeclareMathOperator{\argminH}{argmin}   

\usepackage{wrapfig}
\usepackage{tablefootnote}

\usepackage{algorithm}
\usepackage{algpseudocode}

\usepackage[misc]{ifsym}

\begin{document}
\title{Newer is not always better:
Rethinking transferability metrics, their peculiarities, stability and performance}
\titlerunning{Newer is not always better: Rethinking transferability metrics}
%
\author{Shibal Ibrahim\inst{1}\thanks{This work was completed as an Intern and Student Researcher at Google.}[\Letter] \and
Natalia Ponomareva\inst{2} \and
Rahul Mazumder\inst{1}}
\authorrunning{Shibal Ibrahim, Natalia Ponomareva, \& Rahul Mazumder}
%
\institute{Massachusetts Institute of Technology, Cambridge MA, USA \email{\{shibal,rahulmaz\}@mit.edu} \and
Google Research, New York NY, USA \\
\email{nponomareva@google.com}}
%


\maketitle              
\begin{abstract}
Fine-tuning of large pre-trained image and language models on small customized datasets has become increasingly popular for improved prediction and efficient use of limited resources.
Fine-tuning requires identification of best models to transfer-learn from and quantifying transferability prevents expensive re-training on \textit{all} of the candidate models/tasks pairs.
In this paper, we show that the statistical problems with covariance estimation drive the poor performance of H-score --- a common baseline for newer metrics --- and propose shrinkage-based estimator. This results in up to $80\%$ absolute gain in H-score correlation performance, making it competitive with the state-of-the-art LogME measure. Our  shrinkage-based H-score is $3-10$ times faster to compute compared to LogME. 
Additionally, we look into a less common setting of target (as opposed to source) task selection. We demonstrate previously overlooked problems in such settings with different number of labels, class-imbalance ratios etc. for some recent metrics e.g., NCE, LEEP that resulted in them being misrepresented as leading measures. We propose a correction and recommend measuring correlation performance against relative accuracy in such settings.
We support our findings with $\sim$ 164,000 (fine-tuning trials) experiments on both vision models and graph neural networks.

\end{abstract}

\section{Introduction}
Transfer learning is a set of techniques of using abundant somewhat related source data $p(X^{(s)},Y^{(s)})$ to ensure that a model can generalize well to the target domain, defined as either little amount of labelled data $p(X^{(t)},Y^{(t)})$ (supervised), and/or a lot of unlabelled data $p(X^{(t)})$ (unsupervised transfer learning). 
Transfer learning is most commonly achieved either via fine-tuning or co-training. Fine-tuning is a process of adapting a model trained on source data by using target data to do several optimization steps (for example, stochastic gradient descent) that update the model parameters. Co-training on source and target data usually involves reweighting the instances in some way or enforcing domain irrelevance on feature representation layer, such that the model trained on such combined data works well on target data. 
Fine-tuning is becoming increasing popular because large models like ResNet50  \cite{He2015}, BERT \cite{DBLP:journals/corr/abs-1810-04805}
etc. are released by companies and are easily modifiable. Training such large models from scratch is often prohibitively expensive for the end user. 

In this paper, we are primarily interested in effectively measuring transferability before training of the final model begins. Given a source data/model, a \textbf{transferability measure} quantifies how much knowledge of source domain/model is transferable to the target model. Transferability measures are important for various reasons: they allow understanding of relationships between different learning tasks, selection of highly transferable tasks for joint training on source and target domains, selection of optimal pre-trained source models for the relevant target task, prevention of trial procedures attempting to transfer from each source domain and optimal policy learning in reinforcement learning scenarios (e.g. optimal selection of next task to learn by a robot). If a measure is capable of efficiently and accurately measuring transferability across arbitrary tasks, the problem of task transfer learning is greatly simplified by using the measure to search over candidate transfer sources and targets. 

\noindent\textbf{\textit{Contributions}} 
Our contributions are three-fold:
\begin{enumerate}
\item We show that H-score, commonly used as a baseline for newer transferability measures, suffers from instability due to poor estimation of covariance matrices. We propose shrinkage-based estimation of H-score with regularized covariance estimation techniques from statistical literature. We show $80\%$ absolute increase over the original H-score and show superior performance in majority cases against all newer transferability measures across various fine-tuning scenarios.
\item We present a fast implementation of our estimator that is $3-10$ times faster than state-of-the-art LogME measure.
\item  We identify problems with 3 other transferability measures (NCE, LEEP and $\mathcal{N}$LEEP) in target task selection when either the number of target classes or the class imbalance varies across candidate target tasks. We propose measuring correlation against relative target accuracy (instead of vanilla accuracy) in such scenarios.
\end{enumerate}
Our large set of $\sim 164,000$ fine-tuning experiments with vision models and graph convolutional networks on real-world datasets shows usefulness of our proposals.

This paper is organized as follows. Section \ref{section:metrics} describes general fine-tuning regimes and transfer learning tasks.
Section \ref{sec:related-work} discusses transferability measures. Section \ref{section:h-score} addresses shortcomings of the pioneer transferability measure (H-Score) that arise due to unreliable estimation and proposes a new shrinkage-based estimator for the H-Score. In Section \ref{section:leep-nce-probs}, we demonstrate problems with recent NCE, LEEP and $\mathcal{N}$LEEP metrics and propose a way to address them.
Finally, Section \ref{section:experiments} presents a meta study of all metrics. 

\section{Transferability setup}
\label{section:metrics}
We consider the following fine-tuning scenarios based on existing literature. \\
(i) \textit{Source Model Selection (SMS)}: For a particular target data/task, this regime aims to select the ``optimal" source model (or data) to transfer-learn from, from a collection of candidate models/data.\\
(ii) \textit{Target Task Selection (TTS)}: For a particular (source) model, this regime aims to find the most related target data/task.

In addition, we primarily consider two different fine-tuning strategies: \\
(i) \textit{Linear fine-tuning/head only fine-tuning (LFT)}: All layers except for the penultimate layer are frozen. Only the weights of the head classifier are re-trained while fine-tuning.\\
(ii) \textit{Non-linear fine-tuning (NLFT)}: Any layer can be designated as a feature extractor, up to which all the layers are frozen; the subsequent layers include nonlinear transformations and are re-trained along with the head on target data. 

\section{Related Work} 
\label{sec:related-work}
Recent literature in transfer learning has proposed efficient transferability measures.
Inspired by principles in information theory, Negative Conditional Entropy (NCE) \cite{Tran2019} uses pre-trained source model and evaluates conditional entropy between target pseudo labels (source models' assigned labels) and real target labels. Log Expected Empirical Predictor (LEEP) \cite{Nguyen2020} modifies NCE by using soft predictions from the source model.
Both NCE and LEEP do not directly use feature information, hence they are not applicable for layer selection. 
The authors in \cite{Cui2018} propose representing each output class by the mean of all images from that class and computing Earth Mover's distance between the centroids of the source classes and target classes. 

Other works \cite{Bao2019,Li2021,Huang2021,You2021,Deshpande2021} proposed metrics that capture information from both the (learnt) feature representations and the real target labels. These metrics are more appealing as these can be broadly applicable for models that are pre-trained in either supervised or unsupervised fashion. They are also applicable for embedding layer selection. The authors in \cite{Li2021} proposed $\mathcal{N}$LEEP that fits a Gaussian mixture model on the target feature embeddings and computes the LEEP score between the probabalistic assignment of target features to different clusters and the target labels. The authors in \cite{Huang2021} introduced TransRate --- a computationally-friendly surrogate of mutual information (using coding rate) between the target feature embeddings and the target labels.
H-score was proposed by \cite{Bao2019} that takes into account inter-class feature variance and feature redundancy. 
\cite{You2021} proposed LogME that considers an optimization problem rooted in Bayesian statistics to maximize the marginal likelihood under a linear classifier head. \cite{Deshpande2021} introduced LFC to measure in-class similarity of target feature embeddings across samples.

Finally, the authors in \cite{Tan2021} used Optimal Transport to evaluate domain distance, and combined it, via a linear combination, with NCE.
To learn such a measure, a portion of target tasks were set aside, the models were transferred onto these tasks and the results were used to learn the coefficients for the combined Optimal Transport based Conditional Entropy (OTCE) metric. While the resulting metric appears to be superior over other non-composite metrics like H-score and LEEP, it is expensive to compute since it requires finding the appropriate coefficients for the combination.

\section{Improved estimation of H-score}
\label{section:h-score}
\textit{H-score} \cite{Bao2019} is one of the pioneer measures that is often used as a baseline for newer transferability measures, which often demonstrate the improved performance. It characterizes discriminatory strength of feature embedding for classification:
\begin{align}
    \mathrm{H}(f) = \text{tr}({\B \Sigma^{f}}^{-1} \B \Sigma^{z})
\end{align}
where, $d$ is the embedding dimension, $\B f_i = h(\B x_i^{(t)}) \in \R^{d}$ is the target feature embeddings when the feature extractor ($h: \R^p \rightarrow \R^{d}$) from the source model is applied to the target sample $\B x_i^{(t)} \in \R^{p}$, $\B F \in \R^{n_t \times d}$ denotes the corresponding target feature matrix, $Y = Y^{(t)} \in \mathcal{Y} = \{1, \cdots, C\}$ are the target data labels, $\B \Sigma^{f} \in \R^{d \times d}$ denotes the sample feature covariance matrix of $\B f$, $\B z=\E\left[\B f| Y\right] \in \R^{d}$ and $\B Z \in \R^{n_t \times d}$ denotes the corresponding target-conditioned feature matrix, $\B \Sigma^{z} \in \R^{d \times d}$ denotes the sample covariance matrix of $\B z$.  
Intuitively, $\mathrm{H}(f)$ captures the notion that higher inter-class variance and small feature redundancy results in better transferability.

\begin{wrapfigure}[17]{r}{0.37\textwidth}
    \centering
    \includegraphics[width=0.35\textwidth]{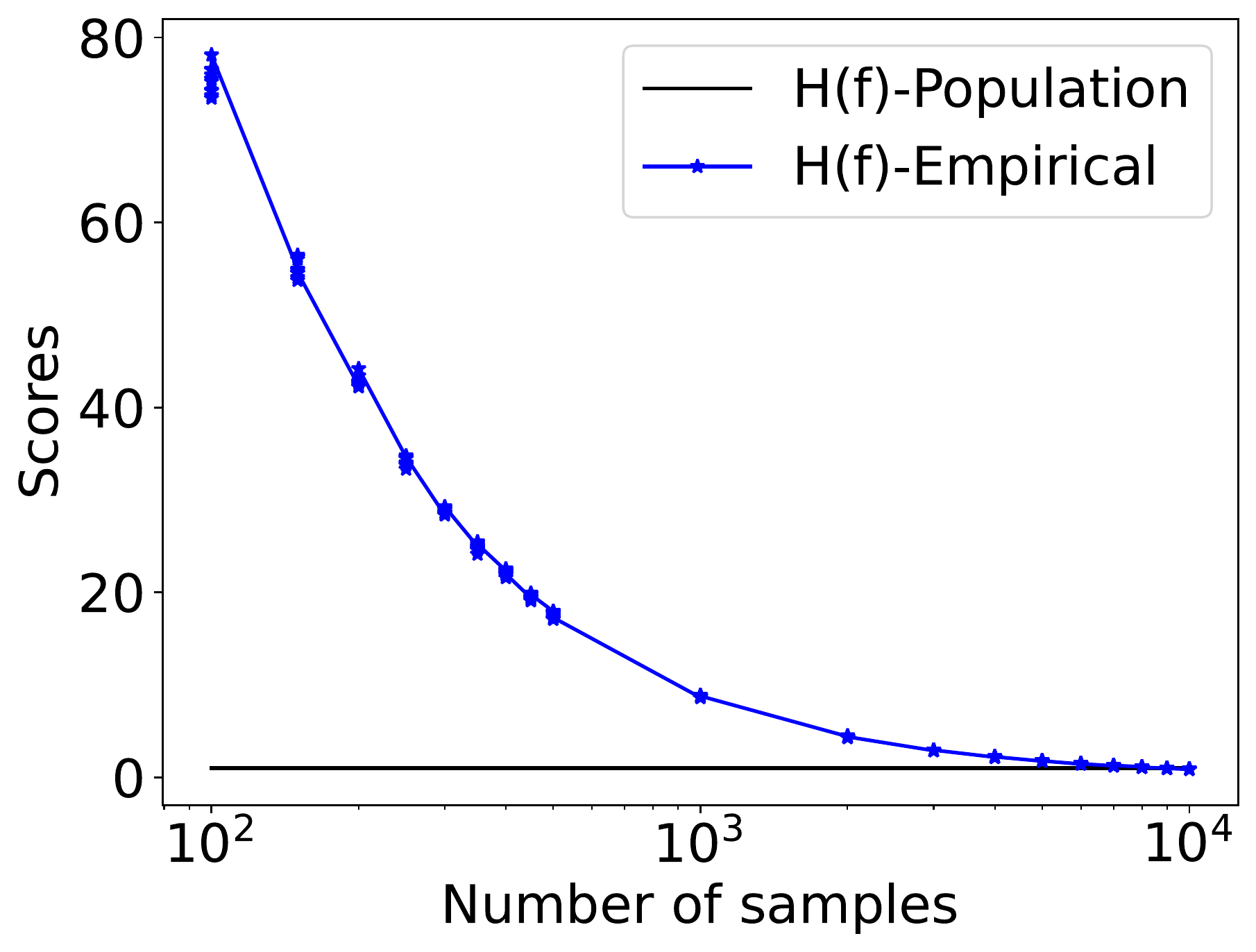}
    \caption{Non-reliability of $\text{H}(f)$. 
    $\text{H}(f)$ is $\sim 75 \times$ larger than the population version of the H-Score. Population version is estimated with $10^6$ samples. 
    }
    \label{fig:hscore_toyexample}
\end{wrapfigure}
We hypothesize that the sub-optimal performance of H-Score (compared to that of more recent metrics) for measuring transferability in many of the evaluation cases, e.g., in \cite{Nguyen2020}, is due to lack of robust estimation of H-Score. We empirically validate this hypothesis using a synthetic classification data. We generated 1 million 1000-dimensional features with 10 classes using Sklearn multi-class dataset generation function \cite{Pedregosa2011}. Number of informative features is set to 500 with rest filled with random noise. We visualize the original and the population version of the H-score for different sample sizes in Fig. \ref{fig:hscore_toyexample}. We observe that the original H-Score becomes highly unreliable as the number of samples decreases. 

Many of the deep learning models in the context of transfer learning have high-dimensional feature embedding space --- typically larger than the number of target samples. Consequently, the estimation of the two covariance matrices in H-score becomes challenging: the sample covariance matrix of the feature embedding has a large condition number\footnote{Condition number of a positive semidefinite matrix $A$, is the ratio of its largest and smallest eigenvalues.} in small data regimes. In many cases, it cannot even be inverted. \cite{Bao2019} used a pseudo-inverse of the covariance matrix $\B \Sigma^{f}$. However, this method of estimating a precision matrix can be sub-optimal as inversion can amplify estimation error \cite{Ledoit2004}. We propose to use well-conditioned shrinkage estimators motivated by the rich literature in statistics on the estimation of  
high-dimensional covariance (and precision) matrices~\cite{pourahmadi2013high}. We show that the use of such shrinkage estimators can offer significant gain in the performance of H-score in predicting transferability. In many cases, as our experiments show, the gain is so significant that H-score becomes a leading transferability measure, surpassing the performance of state-of-the-art measures. 

\subsection{Proposed Transferability Measure} We propose the following shrinkage based H-score: 
\begin{align}
    \label{eq:shrunk-h-score}
    \mathrm{H}_{\alpha}(f) = \text{tr}\bigl( \B \Sigma_{\alpha}^{{f}^{-1}} \cdot (1-\alpha)\B \Sigma^{z}\bigr),
\end{align}
\noindent\textbf{\textit{Estimating $\B \Sigma^{(\B f)}_{\B \alpha}$}} While there are several possibilities to obtain a regularized covariance matrix~\cite{pourahmadi2013high}, we present an approach that considers a linear operation on the eigenvalues of the sample version of the feature embedding covariance matrix. Similar ideas of using well-conditioned plug-in covariance matrices are used in the context of discriminant analysis~\cite{Hastie2001}. In particular, we improve the conditioning of the covariance matrix by considering its weighted convex combination with a scalar multiple of the identity matrix:
\begin{align}
\label{eq:shrinkage-covariance}
\B \Sigma_{\alpha}^{f} = (1 - \alpha)\B \Sigma^{f} + \alpha \sigma \B I_d
\end{align}
where $\alpha \in [0,1]$ is the shrinkage parameter and $\sigma$ is the average variance computed as $\text{tr}(\B \Sigma^{f})/d$. The linear operation on the eigenvalues ensures the covariance estimator is positive definite.
Note that the inverse of $\B \Sigma_{\alpha}^{f}$ can be computed for every $\alpha$, by using the eigen-decomposition of $\B \Sigma^{f}$.
The shrinkage parameter controls the bias and variance trade-off; the optimal $\alpha$  needs to be selected. This distribution-free estimator is well-suited for our application as the explicit convex linear combination is easy to compute and makes the covariance estimates well-conditioned and more accurate \cite{Ledoit2004,Chen2010,Schfer2005}.

\noindent\textbf{\textit{Understanding $(\B 1- \B \alpha)\B \Sigma^{\B z}$}} The scaling factor $(1-\alpha)$ can be understood in terms of regularized covariance matrix estimation under a ridge penalty: 
\begin{align}
\label{eq:ridge-shrinkage}
    1/(1+\lambda) \cdot \B \Sigma^{z}= \argminH_{\hat{\B \Sigma}} ||\hat{\B \Sigma} - \B \Sigma^{z}||_2^2 + \lambda ||\hat{\B \Sigma}||_2^2
\end{align}
where $\lambda \geq 0$ is the ridge penalty. Choosing $\lambda = \alpha/(1-\alpha)$, it becomes clear that $(1-\alpha)\B \Sigma^{z}$ is the regularized covariance matrix.

\noindent\textbf{\textit{Choice of $\B \alpha$}} 
\cite{Ledoit2004} proposed a covariance matrix estimator that minimizes mean squared error loss between the shrinkage based covariance estimator and the true covariance matrix. The optimization considers the following objective:
\begin{align}
    \text{min}_{\alpha, v}~~\E[||\B\Sigma^*-\B\Sigma||^2] ~~~~~~~\text{s.t.}~~\B\Sigma^*= (1-\alpha)\B\Sigma^{f}+\alpha v I, ~~~\E[\B\Sigma^{f}] = \B\Sigma.
\end{align}
where $\norm{\B A}^2 \eqdef \text{tr}(\B A \B A^T)/d$ for a matrix $\B A \in \R^{d \times d}$. This optimization problem permits a closed-form solution for the optimal shrinkage parameter, which is given by:
\begin{align}
\small
    \alpha^{*}&=\E[||\B \Sigma^{f} - \B \Sigma||^2]\big/\E[||\B \Sigma^{f}-(\text{tr}(\B \Sigma)/d) \cdot \B I_d||^2] \\
    &\simeq \min\{(1/n_t^2) \sum_{i \in [n_t]}\frac{||\B f_i \B f_i^T - \B\Sigma^{f}||^2}{||\B \Sigma^{f}-(\text{tr}(\B \Sigma^{f})/d) \cdot \B I_d||^2}, 1\}. \label {eq:ledoit-wolf-practice}
\end{align}
where \eqref{eq:ledoit-wolf-practice} defines a valid estimator (not dependent on true covariance matrix) for practical use. For proof, we refer the readers to Section 2.1 and 3.3 in \cite{Ledoit2004}. 

\begin{figure}[!h]
\centering
    \begin{subfigure}[t]{0.35\textwidth}
    \centering
    \includegraphics[width=\textwidth]{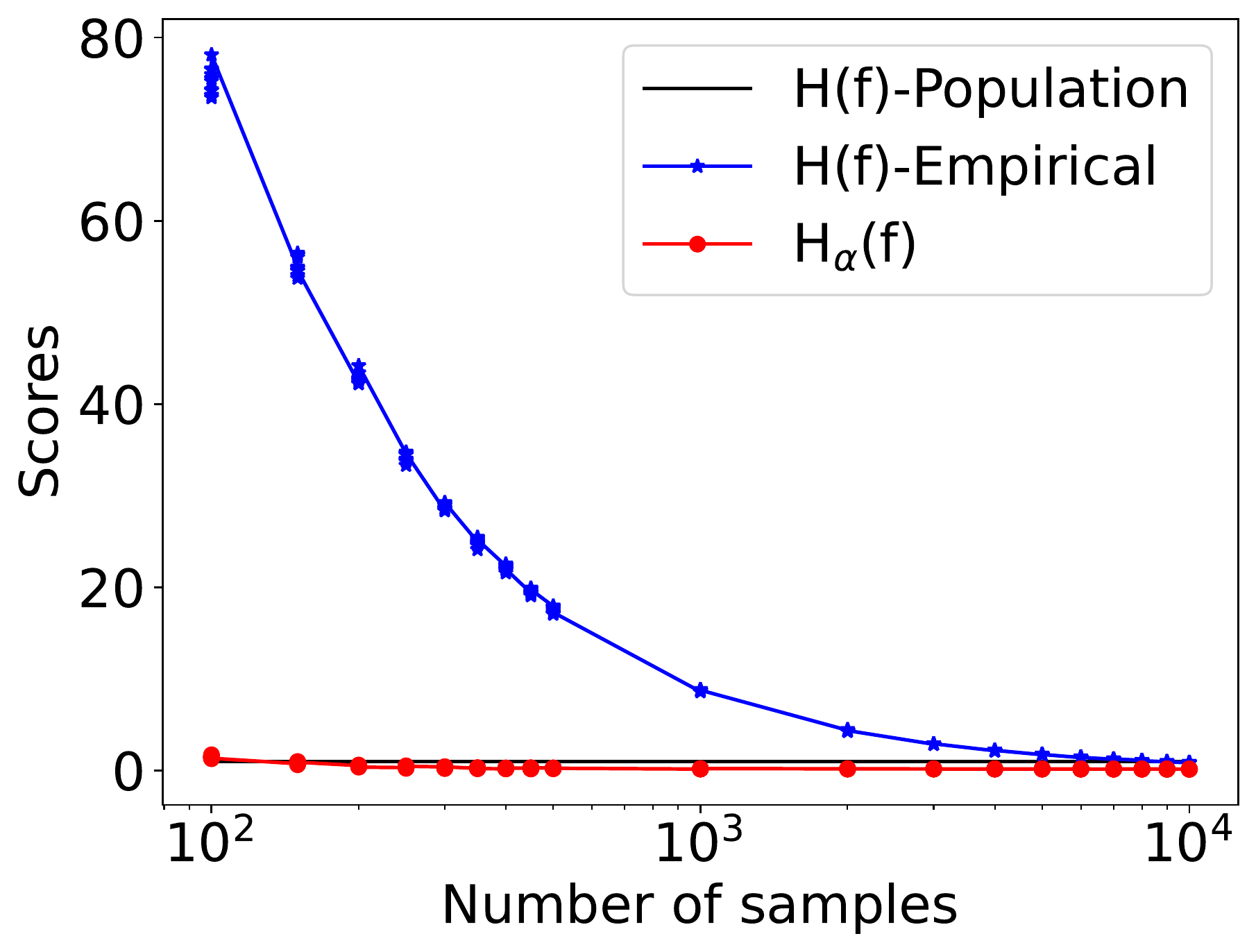}
    \end{subfigure}
    \hspace{0.5cm}
    \begin{subfigure}[t]{0.35\textwidth}
    \centering
    \includegraphics[width=\textwidth]{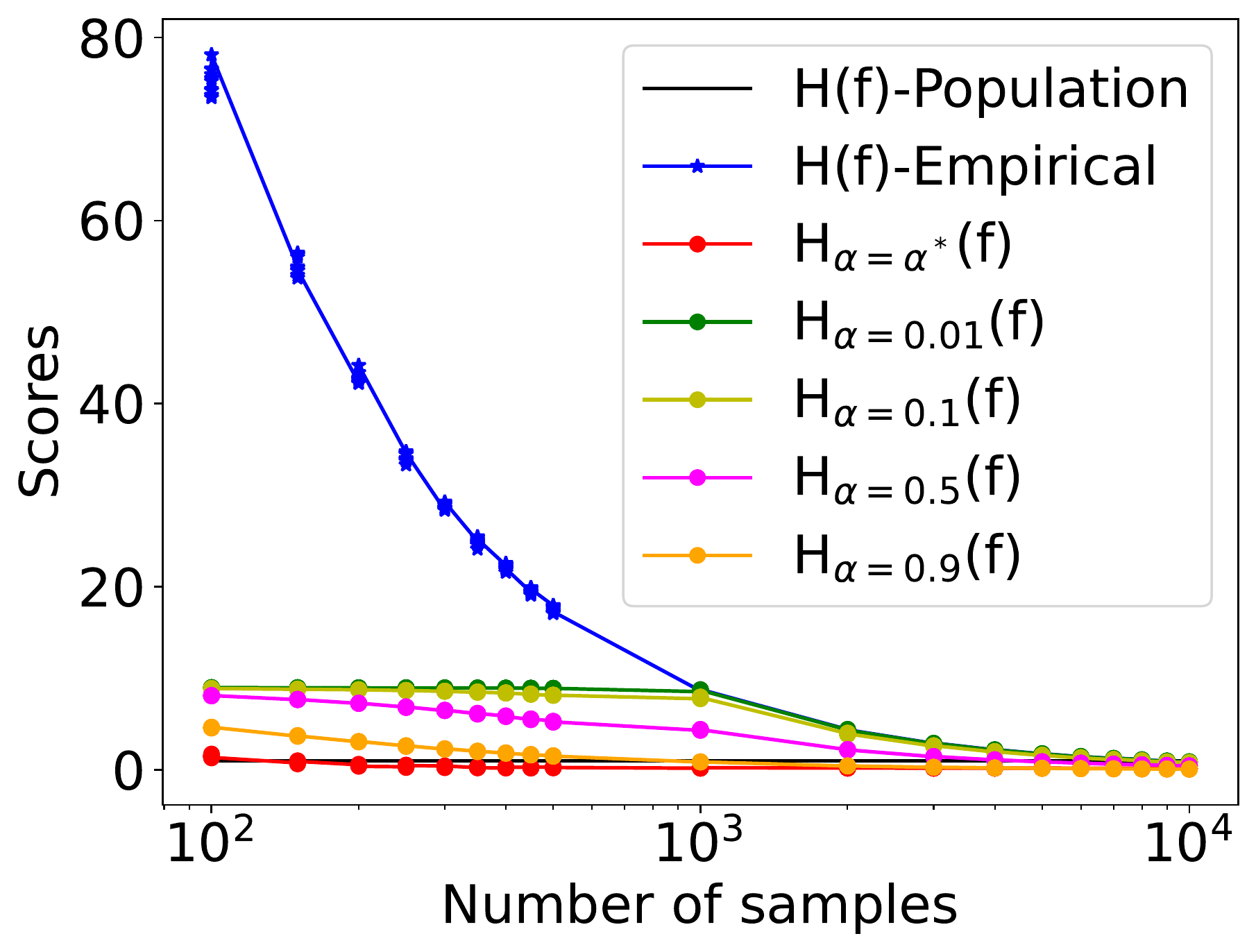}
    \end{subfigure}
    \caption{[Left] Stability of $\text{H}(f)$ and our shrinkage-based $\text{H}_{\alpha}(f)$ with respect to number of samples. $\text{H}(f)$ is $\sim 75$ times larger than the population version of the H-Score (estimated with a sample size of $10^6$). In contrast, the shrinkage-based H-Score is significantly more reliable. [Right] Effect of $\alpha$ on shrunk H-Score.}
    \label{fig:hscore-shrinkage}
\end{figure}

Following the synthetic motivational example presented earlier showing the unreliability of the original H-Score, we investigate the reliability of shrinkage-based H-Score. We visualize the shrinkage-based H-Score in Fig. \ref{fig:hscore-shrinkage}[Left]. We observe that the original H-Score becomes highly unreliable as the number of samples decreases. In contrast, the shrunken estimation of H-Score is highly stable and has a small error when compared with the population H-Score. Hence,  shrinkage-based H-score seems to be a much better estimator of the ``true'' H-score in contrast to the empirical H-Score. We further visualize the effect of using non-optimal values of $\alpha$ on the shrinkage-based H-Score in Fig. \ref{fig:hscore-shrinkage}[Right]. We can see that the shrinkage-based H-Score with optimal shrinkage $\alpha^*$ is much closer to the population version of the original H-Score, especially for smaller sample cases. This validates the use of $\alpha^*$ as computed in \eqref{eq:ledoit-wolf-practice}.

\noindent\textbf{\textit{Additional Discussion on same shrinkage $\B \alpha$ for the two covariances in shrinkage-based H-Score}}
The covariance $\B \Sigma^{z}$ can not be shrunk independently of $\B \Sigma^{f}$ in the estimation of $\mathrm{H}_{\alpha}(f)$--- the two covariances are coupled by the law of total covariance: 
\begin{align}
    \B \Sigma^{f} = \E[\B \Sigma^{f_Y}] + \B \Sigma^{z}. \label{eq:law-of-total-covariance}
\end{align}
\looseness=-1 where $\B f_Y$ denotes the feature embedding of target samples belonging to class $Y \in \mathcal{Y}$ and $\B \Sigma^{f_Y} = \Cov(\B f | Y)$ denotes the class-conditioned covariances. We write
\begin{align}
\small
    (1-\alpha) \B \Sigma^{f} &= (1-\alpha) \E[\B \Sigma^{f_Y}] + (1-\alpha) \B \Sigma^{z}, \nonumber \\
    (1-\alpha) \B \Sigma^{f} + \alpha \frac{\text{tr}(\B \Sigma^{f})}{d}\B I_d &= (1-\alpha) \E[\B \Sigma^{f_Y}] + \alpha \frac{\text{tr}(\B \Sigma^{f})}{d} \B I_d + (1-\alpha) \B \Sigma^{z}, \\
    \text{i.e,}~\B \Sigma_{\alpha}^{f} &= (1-\alpha) \E[\B \Sigma^{f_Y}] + \alpha \frac{\text{tr}(\B \Sigma^{f})}{d} \B I_d + (1-\alpha) \B \Sigma^{z}. \label{eq:shrunk-law-of-total-covariance}
\end{align}
Comparing equations \ref{eq:shrunk-law-of-total-covariance} with \ref{eq:law-of-total-covariance}, we see that the same shrinkage parameter $\alpha$ should be used when using shrinkage estimators, to preserve law of total covariance. 
The first two terms on the right side in \eqref{eq:shrunk-law-of-total-covariance} can be understood as shrinkage of class-conditioned covariances to the average (global) variance. The third term in  \eqref{eq:shrunk-law-of-total-covariance} (e.g. $(1-\alpha)\B \Sigma^{z}$) can then be understood as ridge shrinkage as in \eqref{eq:ridge-shrinkage}.

\subsection{Challenges of comparing $\mathrm{H}_{\alpha}(f)$ across source models/layers}
\label{sec:Challenges of comparing feature distribution discrepancy across task pairs}
Next, we discuss challenges of using $\mathrm{H}_{\alpha}(f)$ on the feature embeddings of target data derived from the source model for source model selection. The feature dimension ($d$) across different source models even for the penultimate layer can vary significantly e.g. from 1024 in MobileNet to 4096 in VGG19. Such differences makes source model/layer selection for fine-tuning highly problematic. 

We propose dimensionality reduction of feature embeddings before computing $\mathrm{H}_{\alpha}(f)$. We project feature embeddings to a lower $q$-dimensional space, where $q$ is taken to be the same across the $K$ candidate models/layers and satisfies: $q \leq \min_{\B f^{(1)}, \B f^{(2)}, \cdots, \B f^{(K)}}{|\B f^{(.)}|}$ where $|.|$ operator denotes the cardinality of the feature spaces. The dimensionality reduction allows for more meaningful comparison of $\mathrm{H}_{\alpha}(f)$ across source/target pairs; this is relevant for source/layer selection. More generally, it also allows for faster and more robust estimate for limited target samples case ($n_t < d$) for linear and nonlinear fine-tuning. In the case of nonlinear fine-tuning, the intermediate layers of visual and language models have really large $d \sim 10^5$, see Table S1 in Supplement
for examples. 

We consider Gaussian Random Projection, which uses a linear transformation matrix $\B V$ to derive the transformed features $\hat{\B F}=\B F \B V$; it samples components from $\mathcal{N}(0, \frac{1}{q})$ to preserve pairwise distances between any two samples of the dataset. Untrained auto-encoders (AE) are other alternatives that have been used to detect covariate shifts in input distributions by  \cite{Rabanser2019}. It is not known how sensitive these untrained AE are to the underlying architecture---using trained AE is less appropriate for use in transferability measurement for fine-tuning as those maybe  more time-consuming than the actual fine-tuning. We demonstrate improved correlation performance of $\mathrm{H}_{\alpha}(f)$ with dimensionality reduction in Table \ref{tab:sms-correlation} in Section \ref{sec:expts-source-model-selection} for source model selection.

\subsection{Efficient Computation for small target data}

For small target data ($C \leq n_t < d$), the naive implementation of $\mathrm{H}_{\alpha}(f)$ can be very slow. We propose an optimized implementation for our shrinkage-based H-Score that exploits diagonal plus low-rank structure of $\B \Sigma_{\alpha}^{(f)}$ for efficient matrix inversion and the low-rank structure of $\B \Sigma^{(z)}$ for faster matrix-matrix multiplications. We assume $\B F$ (and correspondingly $\B Z$) are centered. The optimized computation of $\mathrm{H}_{\alpha}(f)$ is given by: 
\begin{align}
\small
    \mathrm{H}_{\alpha}(f) = (1-\alpha)/(n_t \alpha \sigma)\cdot \left(\norm{\B R}_F^2-(1-\alpha)\cdot\text{vec}\left(\B G\right)^T \text{vec}\left(\B W^{-1}\B G\right)\right), \label{eq:shrunk-h-score-optimized}
\end{align}
where $\B R=\left[\sqrt{n_1}\bar{\B f}_{Y=1}, \cdots, \sqrt{n_C}\bar{\B f}_{Y=C}\right] \in \R^{d \times C}$, $\B G = \B F\B  R \in \R^{n_t \times C}$, $\B W=n_t \alpha \sigma \B I_n +(1-\alpha)\B F \B F^T \in \R^{n_t \times n_t}$. The algorithm (and derivation) is provided in the Supplementary document.
We make a timing comparison of our optimized implementation of $\mathrm{H}_{\alpha}(f)$ against the computational times of the state-of-the-art LogME measure in Section \ref{sec:timing-comparison-between-logme-and-hscore}.

\section{A closer look at NCE, LEEP and $\mathcal{N}$LEEP measures}
\label{section:leep-nce-probs}
\begin{figure}[!b]
\centering
    \begin{subfigure}[t]{0.35\textwidth}
        \centering
        \includegraphics[width=\textwidth]{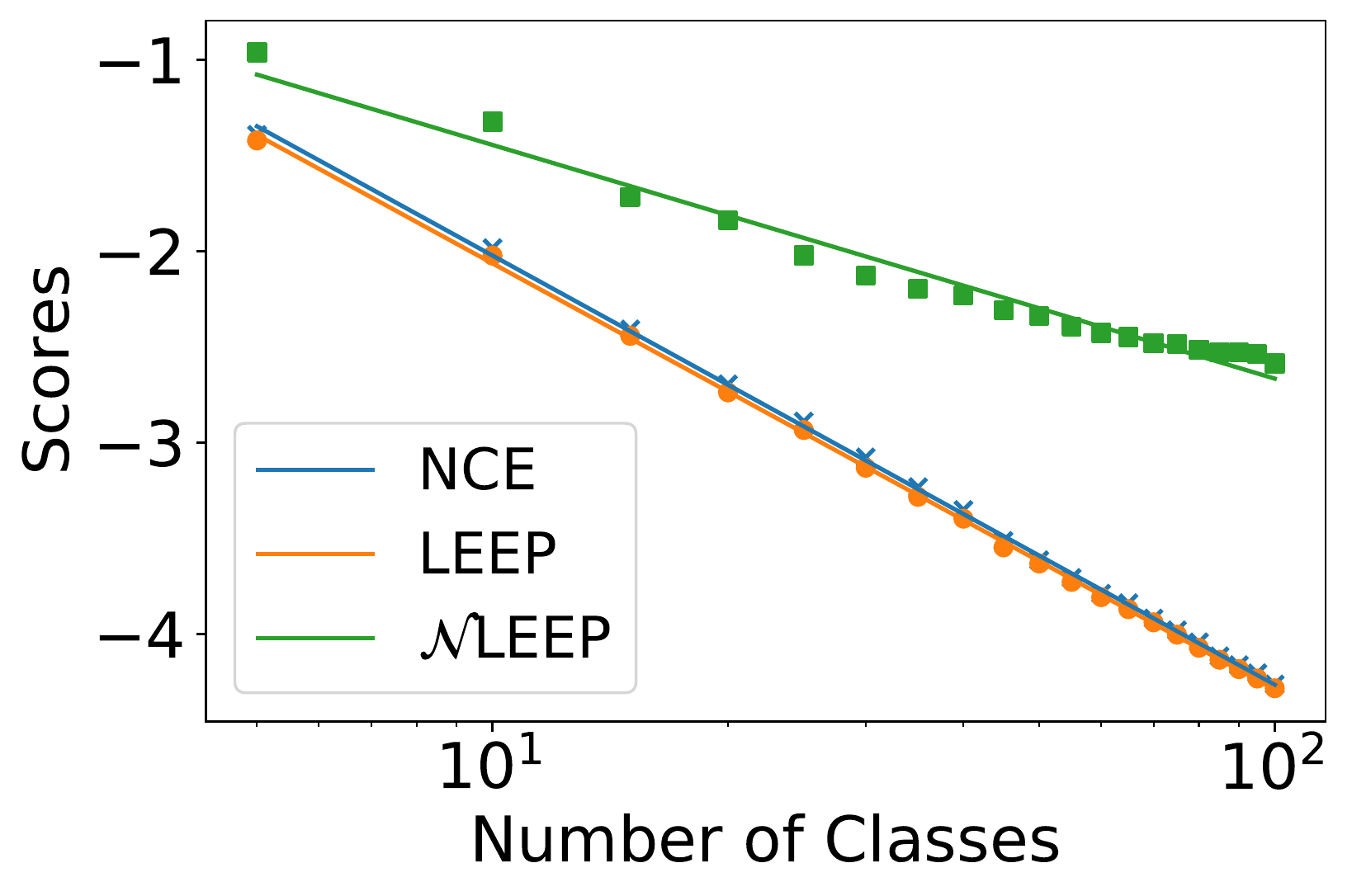}  
    \end{subfigure}%
    \hspace{0.5cm}
    \begin{subfigure}[t]{0.35\textwidth}
        \centering
        \includegraphics[width=\textwidth]{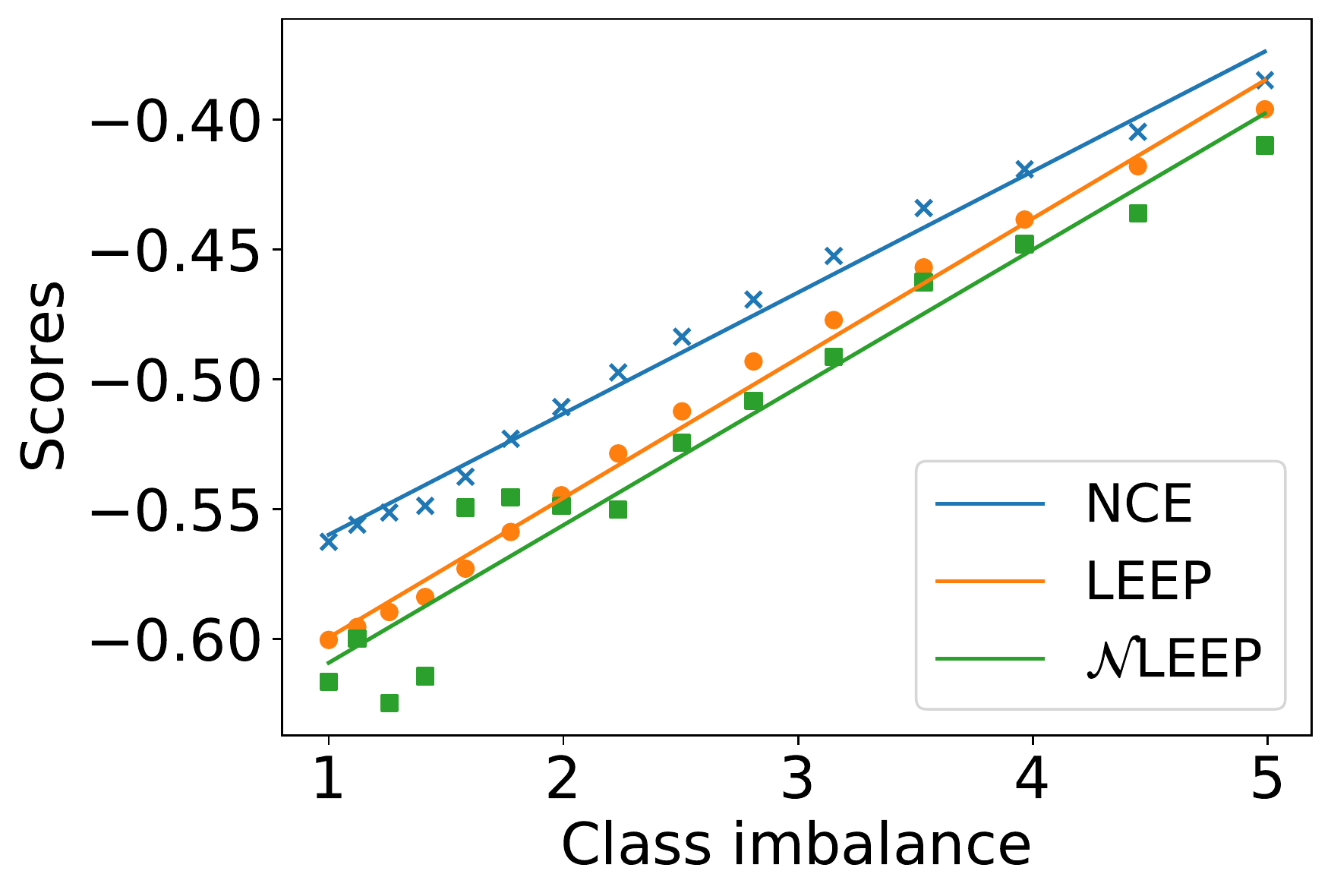}
    \end{subfigure}
    \caption{Relation of NCE, LEEP \& $\mathcal{N}$LEEP to [Left] number of classes (log-scale) and [Right] class imbalance, $max(n_1, n_2)/min(n_1, n_2)$, for VGG19 on CIFAR100. For [Left], we randomly select 2-100 classes. For [Right], we randomly select 2 classes and vary the class imbalances.}
    \label{fig:classes-and-imbalance}
\end{figure}
Next, we pursue a deeper investigation of some of the newer metrics that are reported to be superior to H-Score and bring to light what appears to be some overlooked issues with these metrics in target task selection scenario. Target task selection has received less attention than source model selection. To our knowledge, we are the first to bring to light some problematic aspects with NCE, LEEP  and $\mathcal{N}$LEEP, which can potentially lead to the misuse of these metrics in measuring transferability.

These measures are sensitive to the number of target classes ($C$) and tend to be smaller when $C$ is larger (see Fig. \ref{fig:classes-and-imbalance}[Left]). Therefore, use of these measures for target tasks with \textit{different} $C$ will most likely result in selecting the task with a smaller $C$. 
However, in practice, it is not always the case that transferring to a task with a smaller $C$ is easier; for example, reframing a multiclass classification into a set of binary tasks can create more difficult to learn boundaries \cite{10.1214/aos/1016218223}. Furthermore, the measures are also problematic if two candidate target tasks have different degree of imbalance in their classes even if $C$ is the same. The measures would predict higher transferability for imbalanced data regimes over balanced settings (see Fig. \ref{fig:classes-and-imbalance}[Right]). However, imbalanced datasets are typically harder to learn. 
If these measures are correlated against vanilla accuracy, which tends to be higher as the imbalance increases e.g. for binary classification, the measures would falsely suggest they are good indicators of performance. Earlier work did not consider both these aspects and erroneously showed good correlation of these metrics against vanilla accuracy to show dominance of these metrics in target task selection with different $C$  \cite{Nguyen2020,Tan2021} and imbalance \cite{Tan2021}. 

Here, we propose a method to ameliorate the shortcomings of NCE, LEEP and $\mathcal{N}$LEEP to prevent misuse of these measures, so that they lead to more reliable conclusions.
We propose to standardize the metrics by the entropy of the target label priors, leading to the definitions in \eqref{eq:normalized-NCE-and-LEEP}. This standardization considers both the class imbalance as well as number of classes through the entropy of the target label priors.
\begin{align}
    \text{n-NCE} &\eqdef 1+\text{NCE}/\text{H}(Y), \nonumber \\ 
    ~~~~\text{n-LEEP} &\eqdef 1+\text{LEEP}/\text{H}(Y), \label{eq:normalized-NCE-and-LEEP}\\ 
    ~~~~\text{n-}\mathcal{N}\text{LEEP} &\eqdef 1+\mathcal{N}\text{LEEP}/\text{H}(Y). \nonumber 
\end{align}
Our proposed normalizations in \eqref{eq:normalized-NCE-and-LEEP} ensures the normalized NCE is bounded between $[0,1]$. For proof, see Supplementary document.
n-NCE is in fact equivalent to normalized mutual information and has been extensively used to measure correlation between two different labelings/clustering of samples \cite{Vinh2010}. Given the similar behavior of LEEP and NCE to different $C$ and class imbalance as shown in Fig. \ref{fig:classes-and-imbalance}, we suggest the same normalization as given in \eqref{eq:normalized-NCE-and-LEEP}. However, this normalization does not ensure boundedness of n-LEEP score (and by extension n-$\mathcal{N}$LEEP) in the range $[0,1]$ as in the case of n-NCE. 

For scenarios where candidate target tasks have different $C$, we propose an alternative evaluation criteria (\textit{relative} accuracy) instead of vanilla accuracy --- see Section \ref{section:experiments} for more details. We provide empirical validation of the proposed normalization to these measures in Table \ref{tab:relative-accuracy} in Section \ref{sec:expts-target-task-selection}. We also show that our proposed shrinkage-based H-Score is the leading metric even in these scenarios.

\section{Experiments}
\label{section:experiments}

We evaluate existing transferability measures and our proposed modifications on two class of models: vision models and graph neural networks. We study various fine-tuning regimes and data settings. We draw inspiration from  \cite{Nguyen2020} who consider target task selection and source model selection. The experimental setup highlights important aspects of transferability measures, e.g., dataset size for computing empirical distributions and covariance matrices, number of target classes, and feature dimension etc. Some of these aspects have been overlooked when evaluating transferability measures, leading to improper conclusions.

\noindent\textbf{\textit{Evaluation criteria}} Transferability measures are often evaluated by how well they correlate with the test accuracy after fine-tuning the model on target data. Following \cite{Tran2019,Nguyen2020,Huang2021}, we used Pearson correlation. We include additional results with respect to rank correlations (e.g., Spearman) in Supplementary document.
We argue that considering correlation with the target test accuracy is flawed in some scenarios. In particular, for target task selection, it is wrong to compare target tasks based on accuracy when $C$ is different e.g 5 vs 10 classes. In such a case, task with 10 classes will have a high chance of arriving at lesser test accuracy compared to that for task with 5 classes. In this case, it is more appropriate to consider the gain in accuracy achieved by the model over it's random baseline. Hence we use relative accuracy (for balanced classes): $\frac{\text{Accuracy} - 1/C}{1/C}$.
This measure is more effective in capturing the performance gain achieved by the same model in transferring to two domains with different $C$. This also highlights the limitation of NCE, LEEP and $\mathcal{N}$LEEP which are sensitive to $C$ and tend to have smaller values with higher $C$; these measures do not provide useful information about how hard these different tasks when evaluated with vanilla accuracy.

Correlations marked with asterisks (*) in Tables \ref{tab:shrunk h-score}, \ref{tab:relative-accuracy}, \ref{tab:sms-correlation}, \ref{tab:gcn-tts} 
are not statistically significant ($p$-value $> 0.05$). Larger correlations indicate better identifiability as quantified by transferability measure. Hyphen (-) indicates the computation ran out of memory or was really slow.

\subsection{Case Study: Vision Models}
\label{sec:vision}
First, we evaluate our proposals on visual classification tasks with vision models e.g., VGG19 \cite{Simonyan2015}, ResNet50 \cite{He2015} that have been pre-trained on ImageNet. We fine-tune on subsets of CIFAR-100/CIFAR-10 data. We use Tensorflow Keras \cite{Chollet2015} for our implementation. Imagenet checkpoints come from Keras\protect\footnotemark.
\footnotetext{\url{ https://keras.io/api/applications/}}

\noindent\textbf{\textit{Fine-tuning with hyperparameter optimization}} The optimal choice of hyperparameters for fine-tuning is not only target data dependent but also sensitive to the domain similarity between the source and target datasets \cite{Li2020Finetuning}. We thus tune the hyperparameters for fine-tuning: we use Adam optimizer and tune batch size, learning rate, number of epochs and weight decay (L2 regularization on the classifier head).
For validation tuning, we set aside a portion of the training data (20\%) and try 100 random draws from hyperparameters' multi-dimensional grid. With this additional tuning complexity, we performed $650 \times 100$ fine-tuning experiments. See additional information and motivation in Supplement.

\subsubsection{Target Task Selection}
\label{sec:expts-target-task-selection}
\begin{table}[!t]
\scriptsize
\centering
\caption{Pearson correlation of transferability measures against fine-tuned target accuracy of vision models in the context of target task selection. We compare our proposed $\mathrm{H}_{\alpha}(f)$ against original $\mathrm{H}(f)$ and state-of-the-art measures.  
}
\label{tab:shrunk h-score}
\setlength{\tabcolsep}{1.5pt}
\begin{tabular}{llll|cccccccc}
\multicolumn{1}{c}{Strategy} & \multicolumn{1}{c}{Target} & \multicolumn{1}{c}{Model} & \multicolumn{1}{c|}{Reg.} & $\mathrm{H}(f)$ & $\mathrm{H}_{\alpha}(f)$ & NCE   & LEEP & $\mathcal{\B N}$LEEP  & TransRate & LFC  & LogME \\ \hline
\multicolumn{1}{c}{\multirow{8}{*}{LFT}} & \multirow{4}{*}{CIFAR-100} & \multirow{2}{*}{VGG19}    & S-B  & -0.14* & 0.81 & 0.67 & 0.65 & 0.81 & 0.56 & 0.76 & \textbf{0.85} \\  
                        &                           &                            & S-IB & 0.03* & \textbf{0.77} & 0.57 & 0.63 & 0.70 & 0.46 & 0.47 & 0.75 \\ 
                        &                           & \multirow{2}{*}{ResNet50}  & S-B  & 0.03* & \textbf{0.87} & 0.66 & 0.68 & 0.81 & 0.27 & 0.77 & 0.83 \\ 
                        &                           &                            & S-IB & -0.10 & 0.79 & 0.56 & 0.57 & 0.70 & 0.44 & 0.52 & \textbf{0.82} \\ 
                        & \multirow{4}{*}{CIFAR-10}  & \multirow{2}{*}{VGG19}    & S-B  & 0.00* & 0.67 & 0.52 & 0.60 & 0.61 & 0.42  & 0.44 & \textbf{0.74} \\ 
                        &                           &                            & S-IB & 0.09* & 0.81 & 0.75 & 0.82 & 0.83 & 0.29 & 0.32 & \textbf{0.89} \\ 
                        &              & \multirow{2}{*}{ResNet50}               & S-B  & -0.29 & \textbf{0.73} & 0.43 & 0.44 & 0.61 & -0.02* & 0.57 & 0.71 \\ 
                        &                           &                            & S-IB & 0.17* & \textbf{0.89} & 0.66 & 0.71 & 0.75 & 0.28 & 0.01* & 0.83 \\ \hline
\multicolumn{1}{c}{\multirow{2}{*}{NLFT}} & \multirow{2}{*}{CIFAR-100} & \multirow{2}{*}{VGG19} & S-B  & 0.17* & \textbf{0.73} & 0.58 & 0.59 & 0.67 & -0.03* & 0.70 & - \\ 
                        &                          &                            & S-IB & 0.03* & 0.49 & 0.49 & 0.54 & \textbf{0.55} & 0.48 & 0.17 & - \\ \hline
\end{tabular}
\end{table}
We first investigate the correlation performance of our proposed estimator and existing transferability measures in the context of target task selection. We consider two small target data cases. We provide a summary of the different cases (inspired from \cite{Nguyen2020}) below. Additional details are in the Supplementary document.
\begin{itemize}
    \item \textit{Small-Balanced Target Data (S-B):} We make a random selection of 5 classes from CIFAR-100/CIFAR-10 and sample 50 samples per class from the original train split. We repeat this exercise 50 times (with a different selection of 5 classes). 
    We evaluate correlations of transferability measures across the 50 random experiments. 
    \item \textit{Small-Imbalanced Target Data (S-IB):} We make 50 random selections of 2 classes from CIFAR-100/CIFAR-10, sample between $30-60$ samples from the first class and sample $5 \times$ the number of samples from the second class. This makes for a binary imbalanced classification task. We again measure performance of transferability measures against optimal target test accuracy. Note the imbalance is constant across the candidate target tasks.  
\end{itemize}

\begin{table}[!b]
\scriptsize \centering
\caption{Pearson correlation of transferability measures against \textit{relative} accuracy for large balanced CIFAR-100 dataset with different number of classes across target tasks.}
\label{tab:relative-accuracy}
\setlength{\tabcolsep}{2.5pt}
\begin{tabular}{l|cc|cc|cc|cc|cc}
Model & $\mathrm{H}(f)$ & $\mathrm{H}_{\alpha}(f)$ & NCE & n-NCE & LEEP & n-LEEP & $\mathcal{\B N}$LEEP & n-$\mathcal{\B N}$LEEP  & TransRate & LogME \\ \hline
VGG19          & 0.88 & \textbf{0.97} & -0.95 & 0.66 & -0.95 &  0.66 & -0.93 & 0.95 & 0.68 & 0.96 \\ 
ResNet50       & 0.95 & \textbf{0.98} & -0.95 & -0.74 & -0.95 & -0.73 & -0.94 & -0.63 & 0.56 & 0.96 \\ \hline
\end{tabular}
\end{table}

We evaluate target task selection for linear and nonlinear fine-tuning under small sample setting. The layers designated as embedding layers for nonlinear fine-tuning of VGG19 and ResNet50 is given in Table S1 in Supplementary document. We empirically compare the shrinkage-based H-score against the original measure by \cite{Bao2019} and the state-of-the-art measures. Table \ref{tab:shrunk h-score} demonstrates $80\%$ absolute gains in correlation performance of $\mathrm{H}_{\alpha}(f)$ over $\mathrm{H}(f)$, making it a leading metric in many cases in small target data regimes.

Next, we study a large-balanced target data setting where the number of classes varies across the target tasks. We construct the target tasks as follows: 
We randomly select 2-100 classes from CIFAR-100 and include all samples from the chosen classes. This constructs a collection of large balanced target tasks with different number of classes (L-B-C). We generate 50 such target datasets. In this setting, we evaluate correlation of transferability measures with \textit{relative} target test accuracy for reasons highlighted in Section \ref{section:leep-nce-probs}. This setting validates that the normalizations, proposed in \eqref{eq:normalized-NCE-and-LEEP} in Section \ref{section:leep-nce-probs}, can improve the correlations of NCE, LEEP and $\mathcal{\B N}$LEEP when evaluated on the more appropriate relative accuracy scale. Table \ref{tab:relative-accuracy} demonstrates how various transferability measures perform on target task selection when the number of target classes \textbf{varies}. $\mathrm{H}_{\alpha}(f)$ dominates the performance for both VGG19 and ResNet50 models, surpassing all transferability measures, closely followed by LogME.

\subsubsection{Source Model Selection}
\label{sec:expts-source-model-selection}
Next, we study source model selection scenario for vision models. We select 9 small to large (pre-trained ImageNet) vision models: VGG19, ResNet50, ResNet101, DenseNet121, DenseNet201, Xception, InceptionV3, MobileNet, EfficientNetB0. We evaluate source model selection for linear and nonlinear fine-tuning under small sample setting. The layers designated as embedding layers for nonlinear fine-tuning of all 9 models is shown in Table S1 in Supplementary document.
We sample 50 images per class from all classes available in the original train split of CIFAR-100/CIFAR-10. We designate 10 samples per class for hyperparameter tuning.

We demonstrate that $\mathrm{H}_{\alpha}(f)$ is a leading metric in source model selection as well. 
Given that the feature dimensions vary significantly across different models in source model selection for both linear and nonlinear finetuning, we apply proposed dimensionality reduction via random projection in Section \ref{sec:Challenges of comparing feature distribution discrepancy across task pairs} for $\mathrm{H}_{\alpha}(f)$ to project feature embeddings to $128$-dimensional space ($q=128$). This allows for more meaningful comparison of H-score across source models. This leads to the gains of proposed $\mathrm{H}_{\alpha}(f)$ in terms of correlation in the context of source model selection as well for small samples as given in Table \ref{tab:sms-correlation}, making it again a leading metric in source model selection.

\begin{table}[!t]
\scriptsize \centering
\caption{Pearson correlation of proposed $\mathrm{H}_{\alpha}(f)$ without/with Random Projection (RP) for fine-tuning in source model selection of vision models in small data regimes.}
\label{tab:sms-correlation}
\setlength{\tabcolsep}{1.5pt}
\begin{tabular}{ll|cccccccc}
\multicolumn{1}{c}{Strategy} & Target & $\mathrm{H}(f)$ & $\mathrm{H}_{\alpha}(f)$[No RP] & $\mathrm{H}_{\alpha}(f)$[RP] & NCE & LEEP & $\mathcal{N}$LEEP & TransRate & LogME \\ \hline
\multicolumn{1}{c}{\multirow{2}{*}{LFT}} & CIFAR-100 & -0.190* & 0.024* & \textbf{0.859} & 0.825 & 0.839 & 0.852 & -0.204* & 0.705 \\  
                        & CIFAR-10  & 0.276* & 0.277* & \textbf{0.939} & 0.938 & 0.936 & 0.938 & 0.311* & 0.923 \\ 
\multicolumn{1}{c}{NLFT}               & CIFAR-100 & -0.108* & 0.125* & 0.879 & 0.967 & 0.976 & \textbf{0.977} & - & - \\ \hline
\end{tabular}
\end{table}

\subsection{Case Study: Graph Neural Networks}
\label{sec:gcn}
We evaluate our proposals on Graph Neural Networks on Twitch Social Networks \cite{Rozemberczki2020,Rozemberczki2021,Rozemberczki2021Twitch}. The datasets are social networks of gamers from the streaming service Twitch, where nodes correspond to Twitch users and links correspond to mutual friendships. There are country-specific sub-networks. We consider six sub-networks: \{DE, ES, FR, RU, PTBR, ENGB\}. Features describe the history of games played and the associated task is binary classification of whether a gamer streams adult content. The country specific graphs share the same node features which means that we can perform transfer learning with these datasets. Additional details about the Twitch Social Networks datasets are included in Supplementary document.

\noindent\textbf{\textit{Transferability Setup}} We consider a two-layered Graph Convolutional Network (GCN) \cite{Kipf2017}. The network takes the following functional form:
\begin{align}
\text{logit}(\B{x}) = \hat{\B{G}}\cdot\text{Dropout}(\text{ReLU}(\hat{\B{G}} \cdot \B{x} \cdot \B{W}^1)) \cdot \B{W}^2,
\end{align}
where $\hat{\B{G}} = \hat{\B{D}}^{1/2}(\B{G+I})\hat{\B{D}}^{1/2}$ denotes the renormalization trick from \cite{Kipf2017} when applied to the graph adjacency matrix $\B{G}\in \R^{m \times m}$, $\hat{D}_{ii}=\sum_j (\B G + \B I_m)_{ij}$ denotes the degree of node $i$, $\B{W}^1 \in \R^{p \times d}$ and $\B{W}^2 \in \R^{d \times C}$ denote the learnable weights for the first and second layer respectively. The mapping from logits to $Y$ can be done by applying a softmax and returning the class with the highest probability.

For studying transferability, we consider the target feature embeddings as: $\B F = h(\B X^{(t)}) =  \hat{\B{G}}\cdot\text{Dropout}(\text{ReLU}(\hat{\B{G}} \cdot \B X^{(t)} \cdot \B{W}^1))$. This creates a linear transfer learning strategy, which is similar to the linear fine-tuning regime studied for vision models in section \ref{sec:vision}. We study target task selection in Section \ref{sec:gcn-tts} and source model selection in Supplementary document.

\noindent\textbf{\textit{Pre-training Implementation}} We use PyTorch Geometric \cite{Fey2019} to setup the pre-training of GCN models. The pre-training uses a country-specific subnetwork and performs training, model selection and testing via a 64\%/16\%/20\% split. The training considers transductive learning in graph networks.
We perform 200 hyperparameter trials that tune over Adam learning rates $[10^{-5}, 10^{-1}]$, batch sizes $\{16,32,64\}$, L2 regularization $[10^{-4}, 1]$, dropout of $0.5$ and maximum $1000$ epochs with early stopping with a patience of $50$.

\noindent\textbf{\textit{Fine-tuning}} For fine-tuning experiments, we recover the target feature embeddings by applying optimal pre-trained source model on a different subnetwork of users.
Given we study linear fine-tuning in Graph Networks, we use GridSearchCV with $\ell_2$-regularized Logistic Regression from sklearn \cite{Pedregosa2011} on (only) target train data to perform (stratified) 5-fold cross-validation. We consider 100 values for L2 regularization in the range $[10^{-5}, 10^{3}]$ on the log scale.
The optimal L2 regularization is used to get the optimal model and the test accuracy is computed to measure correlation of transferability measures.  

\subsubsection{Target Task Selection}
\label{sec:gcn-tts}
We pre-train the GCN model with each of the country-specific sub-network. For each country-specific sub-network, $S \in \{\text{DE, ES, FR}, \\ \text{RU, PTBR, ENGB}\}$, we construct 30 different target tasks. We exclude the specific country on which the source model is pre-trained and use the remaining countries to construct different combinations of networks as targets. For example, if the source model is pre-trained on DE, then the target tasks are given by: $\{\text{ES}, \text{FR}, \text{RU}, \text{PTBR}, \text{ENGB}, (\text{ES,FR}),  (\text{ES,RU}), \cdots, (\text{ES,FR,RU,PTBR,ENGB})\}$. 

We study balanced targets in this regime.
Given that the degree of imbalance varies significantly across different country-specific networks, we collect the largest balanced datasets for each target.
Next, we sample $n_t =1000$ nodes for fine-tuning and allocate the remaining nodes as test samples.
We consider two different embedding sizes for GCN network in this study. We also validate our proposals when we allocate 500 samples for fine-tuning. 

\begin{table}[!t]
\scriptsize
\centering
\caption{Pearson correlation of transferability measures against fine-tuned target accuracy of \textit{Graph Convolutional Networks} in Target Task selection scenario. We compare our proposed $\mathrm{H}_{\alpha}(f)$ against original $\mathrm{H}(f)$ and state-of-the-art measures.}
\label{tab:gcn-tts}
\setlength{\tabcolsep}{3.8pt}
\begin{tabular}{ccc|cccccccc}
$n_t$ & Model & Source & $\mathrm{H}(f)$ & $\mathrm{H}_{\alpha}(f)$\tablefootnote{We empirically observed dimensionality reduction with random projection to improve correlation performance for $\mathrm{H}_{\alpha}(f)$ in this target task selection setting as well. We used $q=128$.} & NCE & LEEP & $\mathcal{\B N}$LEEP & LFC & TransRate & LogME \\ \hline
\multirow{6}{*}{500} & \multirow{6}{*}{GCN-256} & DE & 0.10* & 0.35 & 0.15* & 0.15* & 0.40 & \textbf{0.53} & -0.34 & 0.40 \\  
 &  & ES & 0.24 & 0.41 & 0.48 & \textbf{0.50} & -0.13* & 0.24* & -0.34* & 0.20* \\ 
 &  & FR & 0.57 & \textbf{0.61} & -0.03* & 0.01* & -0.05* & 0.26* & -0.30* & 0.44 \\ 
 &  & RU & 0.11* & \textbf{0.34} & -0.19* & -0.17* & -0.12* & 0.04* & -0.21 & 0.11* \\ 
 &  & PTBR & \textbf{0.37} & 0.24* & 0.16* & 0.16* & -0.13* & -0.02* & -0.05* & 0.15* \\ 
 &  & ENGB & 0.48 & \textbf{0.53} & -0.12* & -0.09* & -0.12* & 0.15* & -0.05* & 0.32* \\ \hline
\multirow{6}{*}{1000} & \multirow{6}{*}{GCN-512} & DE & 0.48 & \textbf{0.71} & 0.29* & 0.44 & -0.14* & 0.45 & 0.17* & 0.59 \\  
 &  & ES & 0.68 & \textbf{0.78} & 0.59 & 0.54 & 0.35* & 0.49 & -0.12* & 0.59 \\  
 &  & FR & 0.59 & \textbf{0.61} & 0.25* & 0.27* & 0.04* & 0.10* & 0.58 & 0.22* \\  
 &  & RU & 0.35 & \textbf{0.44} & -0.12* & 0.08* & -0.25* & 0.14* & -0.07* & 0.27* \\ 
 &  & PTBR & 0.67 & \textbf{0.77} & 0.37 & 0.40 & 0.04* & 0.18* & 0.10 & 0.50 \\ 
 &  & ENGB & \textbf{0.82} & 0.81 & -0.04* & -0.08* & 0.17* & 0.26* & 0.15* & 0.65 \\ \hline
\end{tabular}
\end{table}

We present the Pearson correlation performance of transferability measures against fine-tuned target test accuracy in Table \ref{tab:gcn-tts}. The correlations demonstrate our proposed $\mathrm{H}_{\alpha}(f)$ as the leading metric for target task selection for linear fine-tuning in Graph Neural Networks.
We include additional results for source model selection in Supplementary document.

\subsection{Timing comparison between LogME and $\mathrm{H}_{\alpha}(f)$}
\label{sec:timing-comparison-between-logme-and-hscore}
\begin{wraptable}[12]{r}{0.5\textwidth}
\scriptsize
\centering
\caption{Timing comparison of LogME and our $\mathrm{H}_{\alpha}(f)$. All times are in $ms$.}
\label{tab:timing-comparison}
\setlength{\tabcolsep}{3pt}
    \begin{tabular}{ccc|ccc}
    $n_t$ & $d$ & $|\mathcal{Y}|=C$ & LogME & $ \mathrm{H}(f)$& $ \mathrm{H}_{\alpha}(f)$ \\ \hline
    500 & 500 & 50 & 201 & 123 & \textbf{22} \\ 
    500 & 1000 & 50 & 185 & 376 & \textbf{36} \\ 
    500 & 5000 & 50 & 392 & 9680 & \textbf{373} \\ 
    500 & 1000 & 10 & 111 & 259 & \textbf{33} \\ 
    500 & 1000 & 100 & 268 & 271 & \textbf{36} \\ 
    100 & 1000 & 50 & 72 & 255 & \textbf{16} \\ 
    1000 & 1000 & 50 & 318 & 335 & \textbf{75} \\ \hline
    \end{tabular}
\end{wraptable}
We empirically investigate the computational times of $\mathrm{H}_{\alpha}(f)$ when computed via our optimized implementation in \eqref{eq:shrunk-h-score-optimized}. For this exercise, we generate synthetic multi-class classification data using Sklearn \cite{Pedregosa2011} multi-class dataset generation function that is adapted from \cite{Guyon2003}. We investigate different values for number of samples ($n_t$), feature dimension ($d$) and number of classes ($C$). For data generation, we set number of informative features to be 100
with the rest of the features filled with random noise. For LogME, we use a faster variant proposed by \cite{You2021e}. For a fair comparison, we do not use any dimensionality reduction in the computation of $\mathrm{H}_{\alpha}(f)$.
Table \ref{tab:timing-comparison} demonstrates a significant computational advantage of $\mathrm{H}_{\alpha}(f)$ over LogME. We observe $3-10$ times faster computational times.

\section{Conclusion}
We study transferability measures in the context of fine-tuning. Our contributions are three-fold. First, we show that H-score measure, commonly used as a baseline for newer transferability measures, suffers from instability due to poor estimation of covariance matrices. We propose shrinkage-based estimation of H-score with regularized covariance estimation techniques from statistical literature. We show $80\%$ absolute increase over the original H-score and show superior performance in many cases against all newer transferability measures across various model types, fine-tuning scenarios and data settings. Second, we present a fast implementation of our estimator that provides a $3-10$ times computational advantage over state-of-the-art LogME measure.
Third, we identify problems with 3 other transferability measures (NCE, LEEP and $\mathcal{N}$LEEP) in target task selection (an understudied fine-tuning scenario than source model selection) when either the number of target classes or the class imbalance varies across candidate target tasks. We propose an alternative evaluation scheme that measures correlation against relative target accuracy (instead of vanilla accuracy) in such scenarios.
Our large set of $\sim 164,000$ fine-tuning experiments with multiple vision models and graph neural networks in different regimes demonstrates usefulness of our proposals. We leave it for future work to explore how predictive various transferability measures are for co-training regimes (as opposed to fine-tuning).


%
%
%
\bibliographystyle{splncs04}

\section*{Supplementary Material for paper titled ``Newer is not always better:
Rethinking transferability metrics, their peculiarities, stability and performance''}

This supplemental document contains some additional results not included in the main body of the paper. In particular, we present
\begin{enumerate}[label=S\arabic*]
    \item an algorithm (and derivation) for computing $\mathrm{H}_{\alpha}(f)$. 
    \item Proof for normalization of NCE.
    \item Detailed descriptions of experimental settings for Vision models and additional results.
    \item Detailed descriptions of experimental settings for Graph Neural Networks and additional results.
\end{enumerate}
This supplement is not entirely self-contained, so readers might need to refer to the main paper.

\clearpage
\appendix

\setcounter{table}{0}
\renewcommand{\thetable}{S\arabic{table}}%
\setcounter{figure}{0}
\renewcommand{\thefigure}{S\arabic{figure}}%
\setcounter{equation}{0}
\renewcommand{\theequation}{S\arabic{equation}}
\setcounter{section}{0}
\renewcommand{\thesection}{S\arabic{section}}%
\setcounter{footnote}{0}

\section{Optimized Implementation for $\mathrm{H}_{\alpha}(f)$}
\label{supp-sec:derivation-of-optimized-implementation-for-hscore}
\paragraph{Algorithm}
We present our algorithm in Algorithm \ref{algo:shrunk-h-score}. In practice, in steps 2 and 4, we use GaussianRandomProjection and StandardScaler from sklearn respectively. Additionally, we use memory- and compute-optimized sklearn implementation of Ledoit Wolf in step 12 (sklearn.covariance.ledoit\_wolf\_shrinkage)  and step 18 (sklearn.covariance.ledoit\_wolf).

\begin{algorithm}[!h]
\small
\begin{algorithmic}[1]
\Require Data: $\B F, Y$; use\_dimensionality\_reduction, $q$; 
\If{use\_dimensionality\_reduction}
    \State $\B F$ $\gets$ Random-Projection($\B F$, $q$)
\EndIf
\State $\B F$ $\gets$ z-normalize($\B F$) 
\State $n, d \gets \text{size}(\B F)$
\State $\text{number\_of\_samples\_in\_each\_class} \gets \text{unique}(Y)$ 
\State Initialize $\B R \gets \B 0_{d, C} $ 
\For {$\text{c}=1,2,\ldots,C$}
    \State $\B R[:,c] \gets \sqrt{\text{number\_of\_samples\_in\_each\_class}[c]}\bar{\B f}_{Y=c}$
\EndFor
\If{$n<d$}
    \State Compute $\alpha^{*}$ as in equation 7 (main paper).
    \State $\sigma \gets 1$ (as $\B F$ is normalized)
    \State $\B W \gets 
          n\alpha^{*}\sigma\B I_n+(1-\alpha^{*})(\B F \B F^T)$
    \State $\B G = \B F  \B R$
    \State $\mathrm{H}_{\alpha} \gets \frac{1-\alpha^{*}}{n\sigma\alpha^{*}} \left(||\B R||_F^2 - (1-\alpha^*)\left(\text{vec}(\B G)^T \text{vec}(\B W^{-1} \B G)\right)\right)$
\Else
    \State Compute $\alpha^{*}$ and $\B \Sigma_{\alpha^{*}}^{f}$ as in equations 7 and 3 resp. (main paper).
    \State $\mathrm{H}_{\alpha} \gets \frac{1-\alpha^{*}}{n}\text{trace}\left(\left({\B \Sigma_{\alpha^{*}}^{f}}^{-1}\B R\right) \B R^T\right)$
\EndIf
\end{algorithmic}
\caption{Algorithm for computing $\mathrm{H}_{\alpha}(f)$.}
\label{algo:shrunk-h-score}
\end{algorithm}

\paragraph{Derivation} We derive an optimized computation for our proposed shrinkage-based H-score $\mathrm{H}_{\alpha}(f)$ for small target data ($C \leq n_t < d$) as follows:  

\begingroup
\allowdisplaybreaks
\begin{align}
\small
    \mathrm{H}_{\alpha}(f) &= \text{tr}\left( \B \Sigma_{\alpha}^{{f}^{-1}} \cdot (1-\alpha)\B \Sigma^{z}\right) \nonumber \\
    &= \frac{(1-\alpha)}{n_t}\cdot \text{tr}\left(\left(\alpha \sigma \B I_d+\frac{(1-\alpha)}{n_t} \B F^T\B F\right)^{-1}\B  Z^T\B Z\right), \nonumber \\
    &= \frac{(1-\alpha)}{n_t \alpha \sigma}\cdot \text{tr}\left(\left(\B I_d+\frac{(1-\alpha)}{n_t \alpha \sigma} \B F^T\B F\right)^{-1}\B  R\B R^T\right), \nonumber \\ 
    &= \frac{(1-\alpha)}{n_t \alpha \sigma}\cdot \text{tr}\left(\left(\B I_d-(1-\alpha) \cdot  \B F^T\left(n_t \alpha \sigma \B I_n +(1-\alpha)\B F \B F^T\right)^{-1}\B F\right)\B  R\B R^T\right), \label{eq:woodbury} \\
    &= \frac{(1-\alpha)}{n_t \alpha \sigma}\cdot \left(\text{tr}\left(\B R^T\B  R\right)-(1-\alpha)\cdot\text{tr}\left(\B G^T \B W^{-1}\B G\right)\right), \label{eq:trace-property} \\
    &= \frac{(1-\alpha)}{n_t \alpha \sigma}\cdot \left(\norm{\B R}_F^2-(1-\alpha)\cdot\text{vec}\left(\B G\right)^T \text{vec}\left(\B W^{-1}\B G\right)\right), \label{eq:another-trace-property}
\end{align}
\endgroup
where $\B R=\left[\sqrt{n_1}\bar{\B f}_{Y=1}, \cdots, \sqrt{n_C}\bar{\B f}_{Y=C}\right] \in \R^{d \times C}$, $\B G = \B F\B  R \in \R^{n_t \times C}$, $\B W=n_t \alpha \sigma \B I_n +(1-\alpha)\B F \B F^T \in \R^{n_t \times n_t}$, (\ref{eq:woodbury}) follows by Woodbury matrix identity ($(\B I+\B U \B V)^{-1}=\B I-\B U(\B I+\B V \B U)^{-1} \B V$) \cite{Woodbury1950} and (\ref{eq:trace-property}) and (\ref{eq:another-trace-property}) follow by trace properties.

\section{Normalization of NCE}
\label{supp-sec:Normalized NCE and LEEP}
NCE \cite{Tran2019} evaluates conditional entropy between target pseudo labels $Z^{(t)}$ (source model's assigned labels) and actual target labels, as given by:
\begin{equation}
    \text{NCE}(Y^{(t)}|Z^{(t)}) = \frac{1}{n_t}\sum_{i=1}^{n_t}~\text{log}~p_{Y^{(t)}|Z^{(t)}}(y_i|z_i)
\end{equation}
The conditional entropy of the target labels conditioned on the dummy labels (source model' labels on target data) is:
\begin{equation}
    H(Y|Z) = -\sum_{y \in Y, z \in Z} p_{Y,Z}(y,z)~\text{log}~p_{Y|Z}(y|z)
\end{equation}
It holds that $0 \leq H(Y|Z) \leq H(Y)$ (see appendix \ref{supp-sec:Proof of bounds on conditional entropy}). Negative Conditional Entropy (NCE) is given by $\text{NCE} = -H(Y|Z)$ and it holds that $0 \leq -\text{NCE} \leq H(Y)$. We normalize the NCE as follows:
\begin{align}
    0 \geq \frac{\text{NCE}}{H(Y)} \geq -1  ~~~~~\Rightarrow~~~~ 
    0 \leq 1+\frac{\text{NCE}}{H(Y)} \leq 1 \label{eq:normalize by self-entropy}   
\end{align}
where entropy is positive for any practical classification task.

\subsection{Proof of bounds on conditional entropy}
\label{supp-sec:Proof of bounds on conditional entropy}
\begin{align}
    H(Y|Z) &= -\sum_{y \in \mathcal{Y}, z \in \mathcal{Z}} p_{Y,Z}(y,z) \log p_{Y|Z}(y|z) \\
    &= \sum_{z \in \mathcal{Z}}p_{Z}(z)\left[-\sum_{y \in \mathcal{Y}}p_{Y|Z}(y|z) \log p_{Y|Z}(y|z)\right] \\
    &=\sum_{z \in \mathcal{Z}}p_{Z}(z)H(Y|Z=z) \\
    &\geq 0 \label{eq:nonnegativity}
\end{align}
where (\ref{eq:nonnegativity}) holds because $H(Y|Z=z) \geq 0$ and holds with equality if and only if $Y$ is a deterministic function of $Z$.
\begin{align}
    H(Y|Z) &= -\sum_{y \in \mathcal{Y}, z \in \mathcal{Z}} p_{Y,Z}(y,z) \log p_{Y|Z}(y|z) \nonumber \\ 
    &= - \sum_{y \in \mathcal{Y}, z \in \mathcal{Z}} p_{Y,Z}(y,z) \log \frac{p_{Y,Z}(y,z)}{p_{Z}(z)} \nonumber \\
    &= \sum_{y \in \mathcal{Y}, z \in \mathcal{Z}} p_Y(y) \frac{p_{Y,Z}(y,z)}{p_{Y}(y)} \log \frac{p_{Z}(z)}{p_{Y,Z}(y,z)} \nonumber \\
    &\leq \sum_{y \in \mathcal{Y}}p_{Y}(y)~\log \sum_{z \in \mathcal{Z}} \frac{p_{Y,Z}(y,z)}{p_{Y}(y)} \frac{p_{Z}(z)}{p_{Y,Z}(y,z)} ~~~~~\text{(Jensen inequality)} \nonumber \\
    &= - \sum_{y \in \mathcal{Y}} p_{Y}(y) \log p_{Y}(y) = H(Y). \nonumber
\end{align}

\section{Vision Models}
\subsection{Embedding layers for linear and nonlinear Finetuning}
\label{supp-sec:feature-embedding-layers}
\begin{table}[!h]
\scriptsize
\centering
\caption{Feature extraction layer in vision models for nonlinear fine-tuning. The names are from pre-trained ImageNet models in Tensorflow Keras\protect\footnotemark.}
\label{tab:sms-nonlinear}
\setlength{\tabcolsep}{6.0pt}
\begin{tabular}{l|cc|lc}
\hline
\multicolumn{1}{l}{\multirow{2}{*}{}}                      & \multicolumn{2}{|c|}{Linear Fine-Tuning}                                                                  & \multicolumn{2}{c}{Nonlinear Fine-Tuning}                                          \\ \cline{2-5} 
\multicolumn{1}{l}{}                                       & \multicolumn{1}{|c}{Embedding Layer} & Dimension & Embedding Layer & Dimension \\ \hline
\multirow{9}{*}{}  VGG19          & penultimate                                           & 4096                                            & block3\_pool                      & $200704$                            \\ 
                                 ResNet50       & penultimate                                           & 2048                                            & conv2\_block3\_out                & $200704$                            \\ 
                                 ResNet101      & penultimate                                           & 2048                                            & conv2\_block3\_out                & $200704$                            \\ 
                                 DenseNet121    & penultimate                                           & 1024                                            & pool2\_pool                       & $100352$                             \\ 
                                 DenseNet201    & penultimate                                           & 1920                                            & pool2\_pool                       & $100352$                             \\ 
                                 Xception       & penultimate                                           & 2048                                            & add\_6                            & $142688$                            \\  
                                InceptionV3     &  penultimate                                           & 2048                                            & mixed4                            & $110592$                            \\ 
                                  MobileNet      & penultimate                                           & 1024                                            & conv\_pw\_6\_relu                 & $100352$                            \\ 
                                EfficientNetB0 & penultimate                                           & 1280                                            & block3a\_activation               & $112896$                            \\ \hline
\end{tabular}
\end{table}
\footnotetext{\url{ https://keras.io/api/applications/}}

\subsection{Tuning hyperparameters}
\label{supp-sec:tuning}
Current practices for fine-tuning typically involve a selection of values for hyperparameters when retraining the model on target data. Given that the target datasets are typically small in the transfer learning scenarios, the typical strategy is to adopt the default hyperparameters for training large models while using smaller initial learning rate and fewer epochs for fine-tuning. It has been believed that adhering to the original hyperparameters for fine-tuning with small learning rate prevents catastrophic forgetting of the originally learned knowledge or features. Many studies have used fixed hyperparameters (e.g. learning rate, momentum and weight decay, number of epochs) for fine-tuning. However, the choice of hyperparameters is not necessarily optimal for fine-tuning on target tasks. Earlier work has reported that the performance is sensitive to the default hyperparameter selection, in particular learning rate, momentum (for stochastic gradient descent), weight decay and number of epochs \cite{Mahajan2018,Kornblith2019,Li2020Finetuning}. The optimal choice of these parameters is not only target data dependent but also sensitive to the domain similarity between the source and target datasets \cite{Li2020Finetuning}. Therefore, in order to ensure the target task accuracy (against which the correlation of transferability metrics is measured) is optimal, we repeat the fine-tuning exercise for 100 trials of hyperparameter settings. We employ Adam for fine-tuning experiments and optimize over batch size, learning rate, number of epochs and weight decay (L2 regularization on the classifier head). We select the space of these hyperparameters based on existing literature on fine-tuning, e.g. the learning rate is varied in the range $[10^{-5}, 10^{-1}]$, the number of epochs between $\{25,50,75,100,125,150,175,200\}$, the batch size between $\{32, 64, 128\}$ and the weight decay in the range $[10^{-6}, 10^{-2}]$.

\subsection{Target Task Selection Experiments setup}
\label{supp-sec:target_selection}
This evaluation regime is motivated by task transfer policy learning in robotics/ reinforcement learning. Under this regime, transferability measures can be used to greedily optimize a task transfer policy given a collection of tasks. For instance, a robot has to automatically select which new object to pick up. Given that the robot has learned to pick up a few objects before, it would be beneficial for the robot to optimally select the most transferable source/task object pair and improve it's maneuvering ability throughout the process in a highly efficient manner. TMs can also shed light on the relatedness of different tasks in reinforcement learning setups for better understanding. 

We currently evaluate target task selection regime on visual classification tasks with both VGG19 \cite{Simonyan2015} and ResNet50 \cite{He2015} models on subsets of CIFAR-100/CIFAR-10 data under three different dataset regimes following \cite{Nguyen2020}. In all 3 cases outlined below, $20\%$ of the samples from the randomly generated subsets is designated as validation set for hyperparameter tuning to find the model with with optimal validation accuracy. We use all examples in the original test set for evaluating out-of-sample accuracy performance on the target data. \textit{Both} training and validation samples in the subsets are used for computation of transferability metrics and we report the correlation of these measures against the (relative) test accuracies for the randomly generated subsets.
\begin{itemize}[noitemsep,topsep=0pt,parsep=0pt,partopsep=0pt, leftmargin=*]
    \item \textit{Small-Balanced Target Data:} We make a random selection of 5 classes from CIFAR-100/CIFAR-10 and sample 50 samples per class from the original train split, out of which we designate 10 samples per class for validation. We repeat this exercise 50 times (with a different selection of 5 classes), fine-tune the model for each selection (100 hyperparameter tuning trials per selection to find finetuned model with optimal validation accuracy) and evaluate performance of those optimal models in terms of test accuracy. We then evaluate rank correlations of TMs across the 50 experiments with random selection of 5 sub-classes. 
    \item \textit{Small-Imbalanced Target Data:} We make 50 random selections of 2 classes from CIFAR-100/CIFAR-10, sample between $30-60$ samples from the first class and sample $5 \times$ the number of samples from the second class. This makes for a binary imbalanced classification task. We again measure performance of transferability measures against optimal target test accuracy.  
    \item \textit{Large-Balanced Target Data with different number of classes:} We randomly select 2-100 classes from CIFAR-100 and include all samples from the chosen classes (500 samples per class). This constructs a range of large balanced dataset target task selection cases. We evaluate correlation of TMs with relative target test accuracy across the variable number of target classes.
\end{itemize}

\subsection{Spearman Rank Correlations}
\label{supp-sec:rank-correlation}
\begin{table*}[!b]
\scriptsize \centering
\caption{Spearman correlation comparison of transferability measures. Larger correlations indicate better identifiability as quantified by transferability measure. We compared our proposed $\mathrm{H}_{\alpha}(f)$ against original $\mathrm{H}(f)$ and state-of-the-art measures. We evaluate against \textit{relative} accuracy.  
}
\label{tab:spearman-correlation}
\setlength{\tabcolsep}{0.5pt}
\begin{tabular}{lllcl|ccccccc}
\multicolumn{1}{c}{Scen.} & \multicolumn{1}{c}{Strategy} & \multicolumn{1}{c}{Target} & \multicolumn{1}{c}{Model} & \multicolumn{1}{c|}{Reg.} & $\mathrm{H}(f)$ & $\mathrm{H}_{\alpha}(f)$ & n-NCE   & n-LEEP & n-$\mathcal{\B N}$LEEP  & TransRate & LogME \\ \hline
\multirow{10}{*}{TTS} & \multirow{8}{*}{LFT} & \multirow{4}{*}{CIFAR-100} & \multirow{2}{*}{VGG19}    & S-B  & -0.19* & 0.77 & 0.67 & 0.67 & 0.81 & 0.56 & \textbf{0.86} \\  
                        &  &                           &                            & S-IB & -0.07* & 0.71 & 0.64 & 0.63 & 0.72 & 0.40 &  \textbf{0.79} \\ 
                        &  &                           &                            & L-B-C & 0.96  & \textbf{0.97} & 0.50 & 0.44 & 0.95 & 0.91 & 0.96 \\ 
                        & &                           & \multirow{2}{*}{ResNet50}  & S-B  & 0.13* & 0.80 & 0.63 & 0.65 & 0.78 & 0.19* &  \textbf{0.82} \\ 
                        & &                           &                            & S-IB & -0.10* & 0.76 & 0.57 & 0.58 & 0.69 & 0.41 & \textbf{0.81} \\ 
                        & &                           &                            & L-B & 0.98 & \textbf{1.00} & -0.89 & -0.86 & -0.74 & 0.90 & 0.99 \\ 
                        & & \multirow{4}{*}{CIFAR-10}  & \multirow{2}{*}{VGG19}    & S-B  & 0.06* & 0.57 & 0.49 & 0.49 & 0.55 & 0.30  & \textbf{0.65} \\ 
                        & &                           &                            & S-IB & 0.21* & 0.72 & 0.76 & 0.85 & 0.85 & 0.32 & \textbf{0.86} \\  
                        & &              & \multirow{2}{*}{ResNet50}               & S-B  & -0.31 & \textbf{0.60} & 0.28 & 0.29 & 0.51 & 0.03* & 0.59 \\ 
                        & &                           &                            & S-IB & 0.35 & \textbf{0.76} & 0.64 & 0.69 & 0.72 & 0.25* & \textbf{0.76} \\ \cline{2-12} 
& \multirow{2}{*}{NLFT} & \multirow{2}{*}{CIFAR-100} & \multirow{2}{*}{VGG19} & S-B  & -0.00*  & \textbf{0.76} & 0.61 & 0.62 & 0.71 & - & - \\ 
                        & &                          &                            & S-IB & 0.03* & 0.59 & 0.62 & 0.62 & \textbf{0.68} & - & - \\ \hline
\multirow{3}{*}{SMS} & \multirow{2}{*}{LFT} & CIFAR-100 & -    & A-C  & 0.30* & \textbf{0.88} & 0.83 & 0.83 & 0.80 & 0.35* & 0.83 \\  
 &  & CIFAR-10 & -    & A-C  & 0.07* & 0.88 & 0.93 & 0.92 & 0.92 & 0.07* & \textbf{0.95} \\ \cline{2-12}  
 & NLFT & CIFAR-100 & -    & A-C  & 0.052* & \textbf{0.96} & 0.93 & 0.93 & 0.93  & - & - \\ \hline
\end{tabular}
\end{table*}
We present rank correlation performance of all transferability measures across various fine-tuning scenarios (target task selection and source model selection), fine-tuning strategies (linear and nonlinear) in various data regimes. Table \ref{tab:spearman-correlation} combines setups in Tables 1, 2, and 3 (main paper) and presents \textit{Spearman} correlation performance of $\mathrm{H}_{\alpha}(f)$ against transferability measures. Correlations marked with asterisks (*) are not statistically significant ($p$-value $> 0.05$).  Hyphen (-) indicates the computation ran out of memory on 128GB RAM. 

With respect to Spearman correlations in the table above, our shrinkage-based H-score $\mathrm{H}_{\alpha}(f)$ leads in 7/15 cases and LogME leads in 8/15 cases. In terms of Pearson correlations, $\mathrm{H}_{\alpha}(f)$ leads in 9/15 cases (Tables 1, 2, and 3 (main paper)) and LogME leads in 4/15 cases. Additionally, LogME seems to be intractable with respect to memory and computational speed for nonlinear settings where feature dimension is large ($d \sim 10^5$). Our efficient implementation for $\mathrm{H}_{\alpha}(f)$ provides a $3-10$ times computational advantage over LogME. 

\subsection{Experimental code and type of resources}
\label{supp-sec:Experimental code and type of resources}
We use Tensorflow Keras for our implementation. Imagenet checkpoints (ResNet and VGG) come from Keras https://keras.io/api/applications/. For fine-tuning experiments, we use 2 P100 GPUs per model, 15GB RAM per GPU.

\section{Graph Convolutional Networks}
\subsection{Twitch Social Network Dataset Details}
\label{supp-sec:twitch-social-networks}
Twitch Social Network \cite{Rozemberczki2020,Rozemberczki2021,Rozemberczki2021Twitch} contains social networks of gamers from the streaming service Twitch, where nodes correspond to Twitch users and links correspond to mutual friendships. There are country-specific sub-networks. We consider six sub-networks: {DE, ES, FR, RU, PTBR, ENGB}. Features describe the history of games played and the associated task is binary classification whether a gamer streams adult content. The country specific graphs share the same node features which means that we can perform transfer learning with these datasets.

\begin{table}[!htb]
\label{tab:twitch-social-networks}
\scriptsize
\centering
\caption{Data description for Twitch social network of users with country-specific sub-networks.}
\setlength{\tabcolsep}{4.0pt}
\begin{tabular}{c|ccccc}
Country & \#Features & \#Classes & \#Nodes & Class-imbalance & Max. balanced nodes \\ \hline
DE & 3169 & 2 & 9,498 & 1.53 & 7512 \\
ES & 3169 & 2 & 4,648 & 2.42 & 2720 \\
FR & 3169 & 2 & 6,549 & 1.70 & 4848 \\
RU & 3169 & 2 & 4,385 & 3.08 & 2150 \\ 
ENGB & 3169 & 2 & 7,126 & 1.20 & 6476 \\
PTBR & 3169 & 2 & 1,912 & 1.89 & 1322 \\ \hline
\end{tabular}
\end{table}

\subsection{Source Model Selection}
\label{sec:gcn-sms}
For source model selection, we pre-train 42 different GCN models with embedding dimensions in the range $\{128,144,160,\cdots,784\}$. For each embedding dimension, we follow the pre-training procedure and fine-tuning exercise outlined in subsections Pre-training Implementation and Fine-tuning in Section 6.2 (main paper).
\begin{table}[!t]
\scriptsize \centering
\caption{Correlation of transferability measures against fine-tuned target accuracy of \textit{Graph Convolutional Networks} in source model selection scenario.}
\label{tab:gcn-sms}
\setlength{\tabcolsep}{5.0pt}
\begin{tabular}{cc|cccccccc}
Source & Target & $\mathrm{H}(f)$ & $\mathrm{H}_{\alpha}(f)$ & NCE & LEEP & $\mathcal{\B N}$LEEP & LFC & TransRate & LogME \\ \hline
DE & ES & 0.34 & 0.43 & 0.39 & 0.49 & 0.19* & 0.21* & 0.30* & \textbf{0.50} \\ 
DE & FR & 0.25 & 0.41 & -0.11* & -0.13* & -0.24 & 0.07* & 0.20* & \textbf{0.49} \\ 
FR & DE & 0.40 & \textbf{0.56} & -0.01* & -0.02 & 0.04* & 0.54 & 0.42 & \textbf{0.56} \\ 
RU & DE & 0.55 & \textbf{0.62} & 0.04* & -0.10* & 0.31 & 0.52 & 0.48 & 0.59 \\ 
PTBR & ES & 0.35 & \textbf{0.45} & 0.05* & -0.03* & 0.11* & 0.41 & 0.26* & 0.41 \\ 
ES & RU & 0.28* & \textbf{0.39} & 0.20* & 0.16* & -0.00* & -0.12* & 0.24 & 0.25* \\ 
RU & ENGB & 0.14* & \textbf{0.34} & -0.26* & -0.22* & -0.08* & 0.11* & 0.21* & 0.18* \\ 
PTBR & RU & 0.25* & \textbf{0.31} & 0.29* & 0.21* & 0.08* & 0.07* & 0.20* & 0.29* \\ 
ENGB & DE & 0.20 & 0.52 & -0.01* & -0.01 & 0.19* & 0.05* & 0.09* & \textbf{0.61} \\ 
ENGB & FR & 0.36 & 0.58 & -0.04* & -0.08* & -0.10* & 0.13* & 0.34 & \textbf{0.60} \\ 
DE & PTBR & \textbf{0.36} & 0.22* & 0.08* & 0.11* & -0.17* & 0.11* & 0.28* & 0.33 \\ 
PTBR & DE & \textbf{0.63} & 0.34 & 0.31* & 0.38 & 0.10* & 0.10* & \textbf{0.63} & 0.45 \\ 
ENGB & FR & \textbf{0.70} & 0.47 & 0.22* & 0.28* & 0.32 & 0.07* & 0.65 & 0.64 \\ 
ES & DE & \textbf{0.35} & 0.23* & 0.22* & 0.16* & 0.10* & 0.06* & \textbf{0.35} & 0.31 \\ 
ES & FR & 0.38 & 0.19 & -0.08* & -0.03 & -0.13* & 0.38 & \textbf{0.44} & 0.41 \\ \hline
\end{tabular}
\end{table}

For measuring transferability via $\mathrm{H}_{\alpha}(f)$, we first project target feature embeddings derived from each source GCN model to 128-dimensional space with random projection. This is important for effective comparison of $\mathrm{H}_{\alpha}(f)$ across source models with different embedding dimensions as outlined in Section 4.2 (main paper).

We present Pearson correlation performance against target tesk accuracy in Table \ref{tab:gcn-sms} for our proposed $\mathrm{H}_{\alpha}(f)$ and all state-of-the-art transferability measures. We present multiple source-target sub-network pairs for a thorough evaluation of performance of all transferability measures.  We again see $\mathrm{H}_{\alpha}(f)$ as a leading metrics in transferability estimation for source model selection.

\subsection{Experimental code and type of resources}
\label{supp-sec:Experimental code and type of resources-gnn}
We use PyTorch Geometric for our implementation. For pre-training experiments, we use 4 CPUs per model, 16GB RAM per CPU.
\end{document}